\documentclass[]{spie}  

\usepackage{amsmath,amsfonts,amssymb}
\usepackage{booktabs}
\usepackage{graphicx}
\usepackage{caption}
\usepackage{subcaption}
\usepackage{multirow}
\usepackage[colorlinks=true, allcolors=blue]{hyperref}

\title{Deep learning-based lung segmentation and automatic regional template in chest X-ray images for pediatric tuberculosis}

\author[1]{Daniel Capellán-Martín}
\author[1]{Juan J. Gómez-Valverde}
\author[2]{Ramon Sanchez‑Jacob}
\author[1]{David Bermejo-Peláez}
\author[1]{Lara García-Delgado}
\author[3]{Elisa López-Varela}
\author[1]{Maria J. Ledesma-Carbayo}
\affil[1]{\small{Biomedical Image Technologies, Universidad Politécnica de Madrid \& CIBER-BBN, Madrid, Spain}}
\affil[2]{\small{Department of Radiology, Children's National Hospital \& George Washington University School of Medicine, Washington, Columbia, USA}}
\affil[3]{\small{ISGlobal, Hospital Clínic, Universitat de Barcelona, Barcelona, Spain \& DTTC, Stellenbosch University, South Africa}}

\authorinfo{Further author information: (Correspondence: DCM and MJLC)\\DCM: E-mail: daniel.capellan@upm.es\\  MJLC: E-mail: mj.ledesma@upm.es}

\begin{document} 
\maketitle


\begin{abstract}
Tuberculosis (TB) is still considered a leading cause of death and a substantial threat to global child health. Both TB infection and disease are curable using antibiotics. However, most children who die of TB are never diagnosed or treated. In clinical practice, experienced physicians assess TB by examining chest X-rays (CXR). Pediatric CXR has specific challenges compared to adult CXR, which makes TB diagnosis in children more difficult. Computer-aided diagnosis systems supported by Artificial Intelligence have shown performance comparable to experienced radiologist TB readings, which could ease mass TB screening and reduce clinical burden. We propose a multi-view deep learning-based solution which, by following a proposed template, aims to automatically regionalize and extract lung and mediastinal regions of interest from pediatric CXR images where key TB findings may be present. Experimental results have shown accurate region extraction, which can be used for further analysis to confirm TB finding presence and severity assessment. Code publicly available at: \href{https://github.com/dani-capellan/pTB_LungRegionExtractor}{https://github.com/dani-capellan/pTB\_LungRegionExtractor}. 
\end{abstract}


\section{INTRODUCTION}
\label{sec:intro}  

Despite being an ancient disease, tuberculosis (TB) remains a leading cause of death and a substantial threat to global child health, with an estimated annual burden of 1 million new pediatric cases worldwide and 250 000 children dying because of this disease \cite{Holmberg2019TuberculosisChildren, BasuRoy2019ChildrenProtection}. Of particular concern are the children under the age of five years that account for the highest mortality and risk of TB progression \cite{Lopez-Varela2015IncidenceMozambique}. TB is caused by a small, aerobic bacterium called \textit{Mycobacterium Tuberculosis} (Mtb), which generally affects the lungs, although other parts of the body can be also affected \cite{Holmberg2019TuberculosisChildren}. Most children who die of TB are never diagnosed or treated. Screening may be useful to identify children with possible TB and refer them for further testing, otherwise they should be considered for preventive treatment \cite{Vonasek2020ScreeningChildren}. Chest X-rays (CXR), along with symptom inquiry, are considered as the best TB screening methods, due to its higher availability and lower cost compared to other imaging techniques \cite{Ubaidi2018TheCare}. In clinical practice, experienced physicians examine CXR for TB. However, this is a subjective, time-consuming process and carries a significant risk of misclassification of other diseases with similar radiological patterns \cite{Brady2017ErrorAvoidable,Degnan2019PerceptualSolutions}. Besides, the diagnosis of TB is more difficult in young children, given the non-specific nature of their symptoms and the less specific radiological manifestation compared to adults \cite{Galli2016PediatricTuberculosis}. The most frequent lesions in pediatric TB are lymphadenopathy, airway compression, air space consolidation, pleural effusion, cavities, miliary patterns and Ghon focus \cite{Garcia-Basteiro2015RadiologicalMozambique,Richter-Joubert2017AssessmentTuberculosis,George2017IntrathoracicRadiography}. Due to the difficulty to evaluate lymphadenopathy, the most relevant sign to diagnose TB with confidence on CXR, the lateral view is usually considered to facilitate diagnosis \cite{Andronikou2012UsefulnessCross-referencing}.

In this context, computer-aided diagnosis (CAD) systems supported by Artificial Intelligence (AI) algorithms can play an important role in the mass screening of TB by analyzing the CXR images. In recent years, several CE-certified and commercially available solutions have shown performance comparable to experienced radiologist readings \cite{Qin2019UsingSystems,Qin2021TuberculosisAlgorithms}. However, the existing methods do not perform well in pediatric patients and only one system (RADIFY - \href{http://www.edai.africa/}{www.edai.africa/}) is currently being designed for children older than 2 years. Additionally, despite its relevance, this field of research has been scarcely tackled  \cite{Candemir2019AX-rays}, showing an urgent need for the development of AI systems for infants and young children (\textless 3 years). The first steps in a typical CAD system include preprocessing and region of interest (ROI) localization to apply further processing and being able to diagnose more accurately the disease. For TB, the target ROI are the lungs and other structures in the mediastinal region. Most of the current algorithms for detecting and segmenting the lungs are trained and evaluated using healthy subjects, which could have an impact on the correct identification of areas affected by pathology.

As a first step to tackle these challenges in this work, we propose a multi-view deep learning (DL)-based approach which aims to automatically extract lung and mediastinal regions of interest from pediatric CXR images where key TB findings may be present. The output of the proposed method is a standardized partition of the pediatric CXR that will enable further development of TB radiological sign detection methods as well as the potential assessment of severity.

\section{METHODOLOGY}
\label{sec:methodology}

\begin{figure}[htbp]
  \centering
  \includegraphics[width=0.6\textwidth]{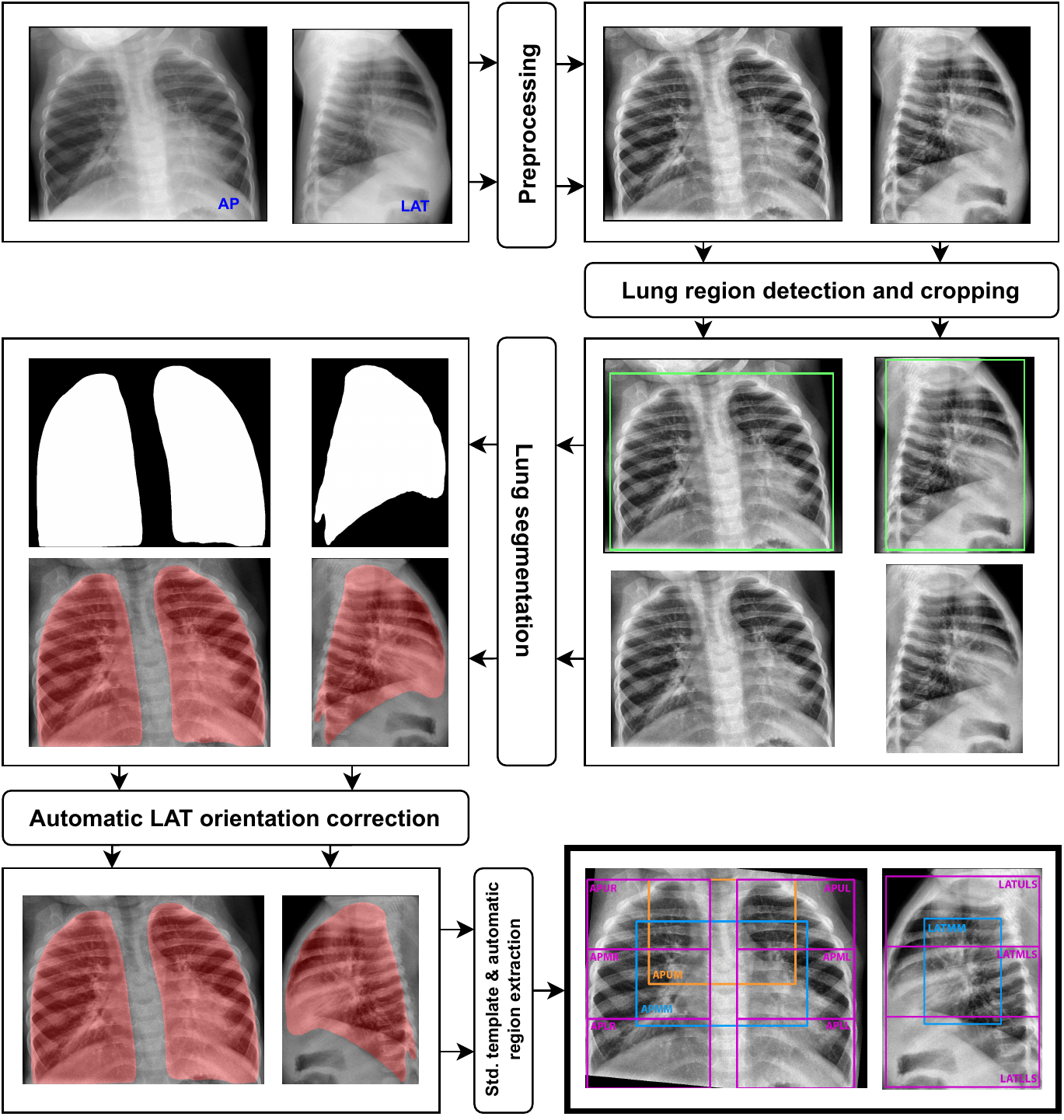}
  \vspace{0.3cm}
  \caption{Pipeline of the proposed solution. Images shown in the pipeline are real predictions and outputs made by the corresponding DL-based models and algorithms on an 8-month-old infant who belongs to the testing set. AP: Anteroposterior. LAT: Lateral.}
  \label{fig:pipeline}
\end{figure}

Figure \ref{fig:pipeline} shows the main steps that make up the proposed solution.

\subsection{Datasets and Splits}

For developing the solution, two datasets were used. Our target CXR dataset is from a pediatric (\textless3 years of age) cohort of 218 TB and non-TB children obtained from a study performed at CISM (Manhiça Health Research Center, Mozambique), with both anteroposterior (AP) and lateral (LAT)-view CXR images \cite{Garcia-Basteiro2015RadiologicalMozambique}. Additionaly, for development we used a subset from the public NIH CXR, ChestX-ray8 dataset (112,120 frontal-view CXR images from 30,805 patients) presenting common thoracic pathologies, such as lung nodules, pneumonia, fibrosis, edema or cardiomegaly \cite{Wang2017ChestX-ray8:Diseases}. To obtain a fully pediatric subset, only images of patients $\leq$11 years old were considered, which account for a total of 2330 images from 847 patients. Since further manual labeling of the images was required, a final subset of 210 images covering different ages and pathological findings was randomly selected.


In the experiments, training and validation splits were considered. The amount of training and validation data is specified later in each of the tasks. To test the proposed solution, an independent CISM subset of 30 patients with both AP and LAT chest X-rays was used.

\subsection{Preprocessing}

To enable comparable contrast representation across the data, a first preprocessing step was applied to the images, mainly based on the application of an image contrast enhancement process with Contrast Limited Adaptive Histogram Equalization (CLAHE), capable of improving local contrast and edges adaptively with modest computational requirements, which has been shown to improve detection of TB and lung lesions on chest radiographs \cite{Zuiderveld1994ContrastEqualization, Lakhani2017DeepSundaram, Siracusano2020Pipelineclahe}. Preprocessing with CLAHE may also provide better performance in image segmentation \cite{Reamaroon2020RobustSyndrome}.


\subsection{Lung Region Detection \& Cropping}
\label{ssec:region-detection}

In the high burden clinical scenarios, digital as well as analog X-ray systems exist. To ensure the same field of view (FOV) and the proper processing of manually digitized X-rays, a lung region detection process was performed to both AP and LAT images. Indeed, first experiments showed that the subsequent lung segmentation process was much more robust when a previous cropping step was included. 

Consequently, two DL-based fully convolutional neural network (FCNN) object detection models, one for AP and another for LAT, based on YOLO (\textit{You Only Look Once}) architecture were implemented. For this, Ultralytics' YOLOv5\footnote{\href{https://github.com/ultralytics/yolov5}{https://github.com/ultralytics/yolov5}} implementation was used for training a lung detector for both AP and LAT images.

For AP images, a YOLOv5s model was trained on a subset of 254 AP images from NIH and CISM datasets (210 and 44, respectively). For LAT images, another YOLOv5s model was trained, this time on 139 LAT images from CISM. All AP and LAT images were manually annotated using CVAT annotation tool and checked by an expert radiologist (RSJ, from author list).

The corresponding object detection outputs were then used to crop both AP and LAT images, narrowing down the field of study to our region of interest, the lungs, thus providing a more robust subsequent segmentation process.

\subsection{Lung Segmentation}

This step gathers one of the most important parts within the pipeline proposed. The lung segmentation was defined to cover all the lung parenchymal extension, independently of the presence of overlapping structures. This is particularly important for pediatric TB cases as some of the findings could appear behind or in front of other structures such as the heart or at the lower posterior lobes of the lungs.

To tackle this, a comparison of three different state-of-the-art DL-based image segmentation architectures was carried out. Different models were trained and tested for each of the views (AP and LAT). Training was performed from scratch. All the data used for this task, including both training and test sets, were manually segmented using annotation tools. These were then checked by an expert radiologist (RSJ, from author list). 

Two U-Net-based architectures and one Transformer-based architecture were used: Gated-Axial Attention UNet (GatedAxialUNet) \cite{Valanarasu2021MedicalSegmentation}, Medical Transformer (MedT) \cite{Valanarasu2021MedicalSegmentation} and nnU-Net ("no-new-Net") \cite{Isensee2020NnU-Net:Segmentation, Ronneberger2015U-Net:Segmentation}. No major changes were made to the source code of each of the implementations, preserving as much as possible default settings. 

In order to assess the performance of each of the models in relation to the amount of supervised data used to train the networks, an incremental learning approach was followed. Supervised training data was progressively increased from 20 to 60 in 20-image steps, gathering segmentation performance results on the independent test set throughout each of the steps.


In the cases of GatedAxialUNet and MedT, input images were resized to an input size of $256\times256$, default batch size of 4 was kept, Adam optimizer was used with the default learning rate value of 0.001, and a total of 400 epochs were considered for training the models. The rest of the hyperparameters and configurations were kept with their default values. The validation set accounted for the 20\% of the initial training set. To train GatedAxialUNet and MedT networks, binary cross-entropy (CE) loss was used between the prediction and the ground truth, which has the following form:

\footnotesize
\begin{equation}
\centering
\mathcal{L}_{CE(p, \hat{p})}=-\left(\frac{1}{w h} \sum_{x=0}^{w-1} \sum_{y=0}^{h-1}(p \log (\hat{p}))+(1-p) \log (1-\hat{p})\right)
\end{equation}
\normalsize


where $w$ and $h$ are the dimensions of the image, $p$, i.e. ($p(x, y)$), corresponds to the pixel label in the image and $\hat{p}$, i.e. $\hat{p}(x, y)$, denotes the output prediction at a specific location $(x,y)$ in the image.

In the case of nnU-Net, 2D U-Net models were trained on the data. Input images were automatically adapted by the implementation, with different patch sizes depending on the image type (AP images: $768\times896$, LAT images: $1024\times896$). The input batch size was automatically set to 1 by the implementation, according to GPU memory requirements. 50 epochs were considered for training the models. Stochastic gradient descent (SGD) with Nesterov momentum ($\mu=0.99$), an initial learning rate of 0.01 and a weight decay following the ‘poly’ learning rate policy, were used for learning network weights. The rest of the hyperparameters and configurations were kept with their default values. Validation sets accounted for 20\% of the initial training set, as a 5-fold cross-validation approach for training the models was adapted, following the implementation's documentation and guidelines. To train the nnU-Net models, a combination of Dice and CE loss was used:

\begin{equation}
\mathcal{L}_{\text {total}}=\mathcal{L}_{\text {Dice}}+\mathcal{L}_{CE}
\end{equation}

where $\mathcal{L}_{CE}$ was defined above and $\mathcal{L}_{Dice}$, for an image $x$ with a binary prediction output $\hat{y}$ and binary label $y$, is defined as:

\begin{equation}
\mathcal{L}_{\text {Dice}}=-\frac{2 \sum_{i} y_{i} \hat{y}_{i}}{\sum_{i} y_{i}+\sum_{i} \hat{y}_{i}}
\end{equation}

where $i$ represents the $i$-th pixel on the images. For further details on the nnU-Net training process, please refer to Isensee et al \cite{Isensee2020NnU-Net:Segmentation}.

\subsection{Automatic LAT Orientation Correction}

In clinical routine, LAT images can be acquired either facing the patient right or left. Depending on this fact, the vertebral column may appear at the right or left side of the image. Consequently, after segmenting the lungs, an automatic orientation correction of the LAT image was included in the pipeline. This provides the solution with robustness and homogeneity, otherwise incorrect regions could be extracted in the subsequent steps.

To tackle this issue, a light-weighted and efficient ResNet-based Deep Convolutional Neural Network was designed and trained from scratch, which learned to detect the vertebrae in the image. The model was trained on 111 CISM LAT images and validated on 28 CISM LAT images (20\% validation split). An horizontal flip was then performed to those images in which the network detected the column at the left (see Figure \ref{fig:pipeline}), homogenizing the data for the automatic region extraction process, and, thus, making the system more robust.

In order to make the training of the network more efficient, input images were first normalized to zero mean and unit variance, using z-normalization ($X_{norm} = \frac{X-\mu}{\sigma+\epsilon}$), where $X$ is the image, $\mu$ is the mean value of the image, $\sigma$ its standard deviation and $\epsilon$ a small ($\epsilon\ll\sigma, \epsilon \approx e^{-10}$) parameter that prevents division by zero.

DL models in this section were implemented using TensorFlow framework. Training and testing in this and previous sections were done using a workstation with NVIDIA TITAN X 12GB and TITAN Xp 12GB GPUs, 64GB RAM and Intel Core i7 @ 3.6 GHz CPU.

\subsection{Standardized Template and Automatic Region Extraction}
\label{ssec:auto_region_extraction_methods}

As a final step, an automatic standardized template, based on Andronikou et al.~proposals \cite{Andronikou2009AdvancesAdults} to regionalize the pediatric CXR, was constructed having as input the previous cropped AP and LAT images, with their corresponding predicted lung segmentations.

To ensure the correspondence of regions across views we first aligned AP and LAT views. Subsequently, the AP image was automatically rotated to ensure lung verticality based on the orientation of the segmentations. To achieve this, a BLOb (Binary Large Object) detection followed by a principal component analysis (PCA) was applied to the AP predicted segmentation masks in order to estimate the rotation of each of the lungs.



The AP and LAT bounding boxes enclosing the lung segmentations were extracted and mediastinal regions were defined based on relative measures with respect to the lungs. The final regions extracted are detailed in Table \ref{tab:regions}. AP and LAT lungs were divided into thirds. LAT lungs were also divided in thirds to identify corresponding areas of potential pathology, not necessarily anatomical regions. APUM contains the respiratory track and suprahiliar area; APMM mainly contains the parahiliar area; and LATMM gathers the parahiliar area, of vital importance for experienced radiologists when detecting parahiliar lymphadenopathies. This standard template and its partitions can be used for further analysis to confirm TB finding presence and severity assessment.

\begin{table}[htb]
\centering
\resizebox{0.6\textwidth}{!}{%
\begin{tabular}{@{}ll@{}}
\toprule
\textbf{View \& Region(s)}                & \textbf{Acronym(s)}      \\ \midrule
AP right lung thirds (upper, middle, lower) & APUR, APMR, APLR       \\
AP left lung thirds (upper, middle, lower)  & APUL, APML, APLL       \\
AP upper and middle mediastinal regions   & APUM, APMM             \\
LAT lungs thirds (upper, middle, lower)     & LATULS, LATMLS, LATLLS \\
LAT middle mediastinal region             & LATMM                  \\ \bottomrule
\end{tabular}%
}
\vspace{0.3cm}
\caption{Extracted regions and their acronyms.}
\label{tab:regions}
\end{table}


\vspace{-0.1cm}
\section{EXPERIMENTS AND RESULTS}
\label{sec:experiments}
\vspace{0.2cm}

\subsection{Lung Region Detection \& Cropping}

Lung detection performance was satisfactory using YOLOv5s, the small version of YOLOv5 (7.2M parameters, 14 MB of size). 

\begin{figure}[htb]
    \centering
    \begin{subfigure}[b]{0.35\textwidth}
        \centering
        \includegraphics[width=\textwidth]{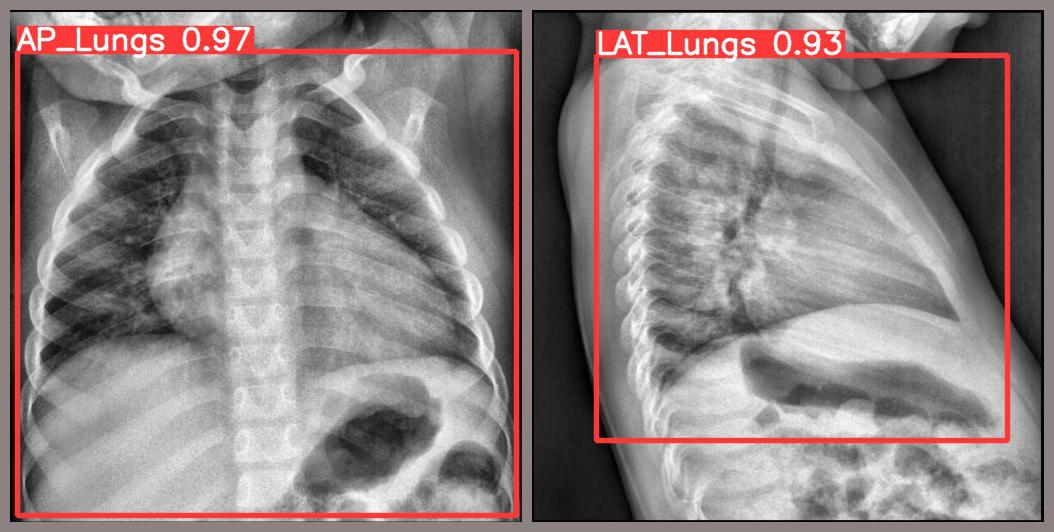}
        \caption{\centering 32-month-old non-TB patient.}
        \label{fig:YOLO_healthy}
    \end{subfigure}
    \hspace{0.1cm}
    \begin{subfigure}[b]{0.35\textwidth}
        \centering
        \includegraphics[width=\textwidth]{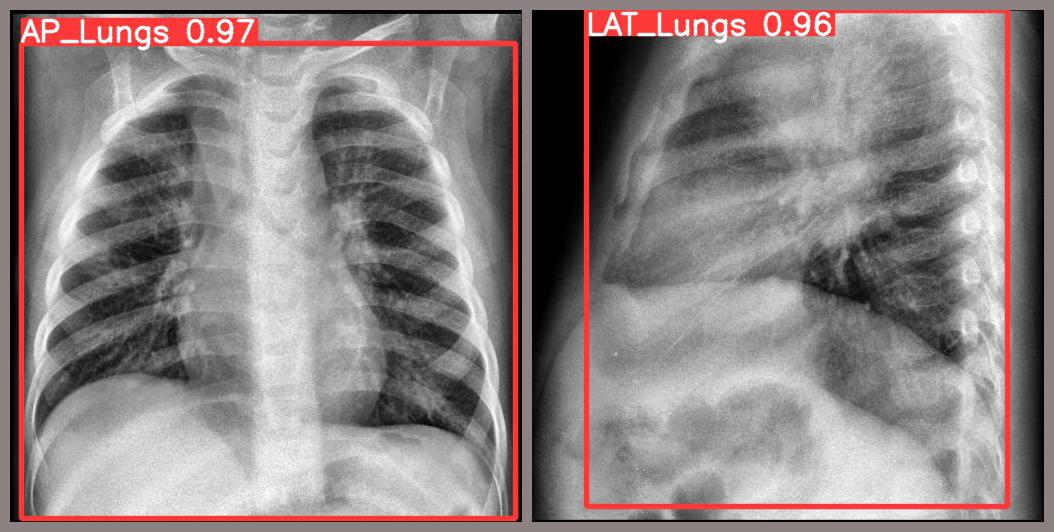}
        \caption{\centering 16-month-old TB patient.}
        \label{fig:YOLO_tb}
    \end{subfigure}
    \vspace{0.3cm}
    \caption{Lung detections in both AP and LAT views of two cases from the test set.}
    \label{fig:YOLO}
\end{figure}

A confidence threshold of 0.7 was selected for inference, with the aim of properly detecting the lungs with these models. Figure \ref{fig:YOLO} shows two examples of how YOLOv5 performs on both AP and LAT views from two testing cases, one non-TB and other TB.

\subsection{Lung Segmentation}

Results obtained throughout all the different experiments carried out in this step are presented in Table \ref{tab:ap-lat-seg-metrics}. These results were obtained by testing the different trained configurations and architectures, following the mentioned incremental learning approach, on an independent CISM test set of 30 manually segmented cases (with their corresponding 30 AP and 30 LAT images). When computing the metrics, all predicted and reference masks were resized to $256\times256$, avoiding metric miscalculation due to this fact.

\subsubsection{Incremental learning}

\begin{figure}[htb]
    \centering
    \begin{subfigure}[b]{0.38\textwidth}
        \centering
        \includegraphics[width=\textwidth]{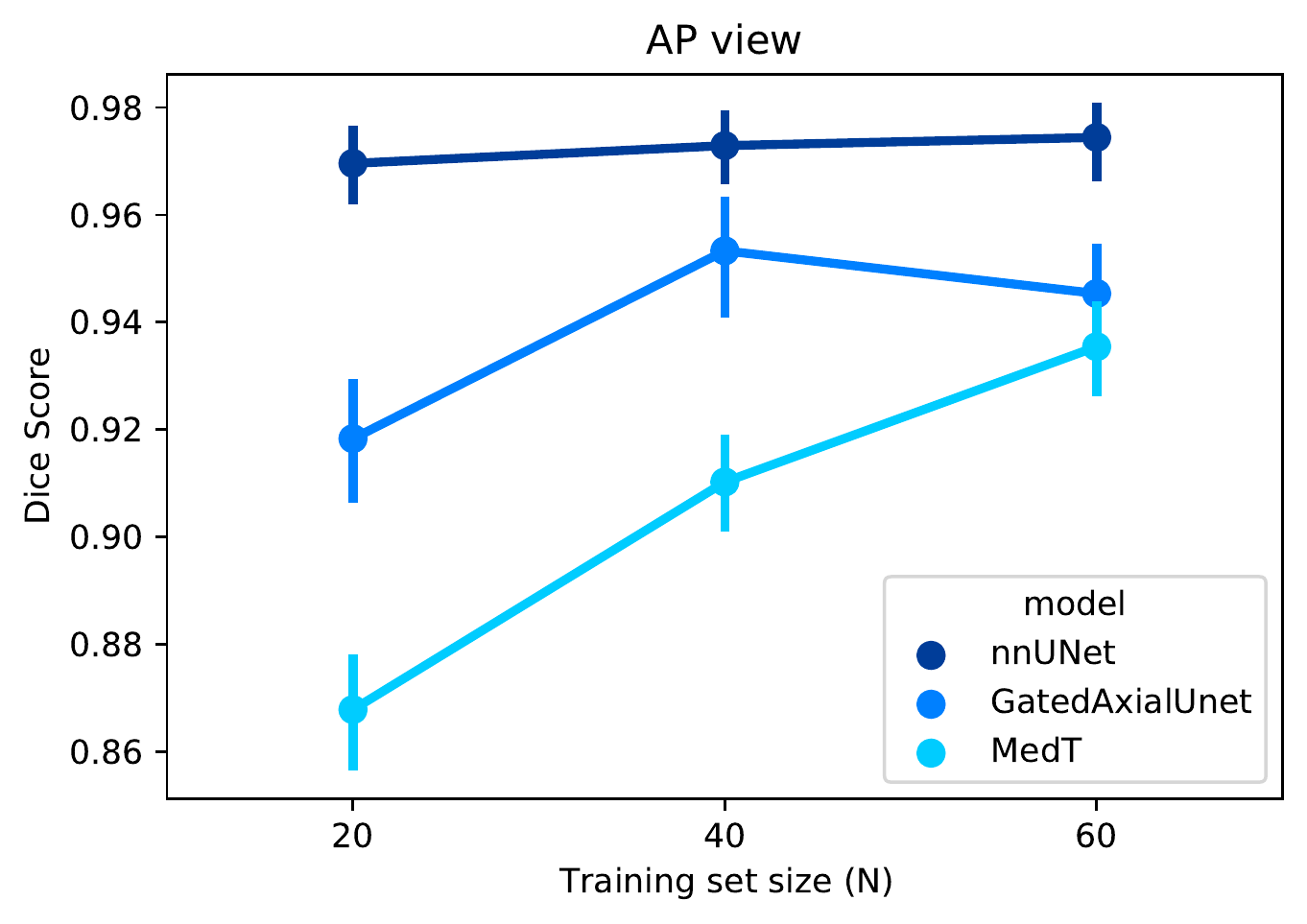}
        \caption{AP Dice}
        \label{fig:AP_dice}
    \end{subfigure}
    \begin{subfigure}[b]{0.38\textwidth}
        \centering
        \includegraphics[width=\textwidth]{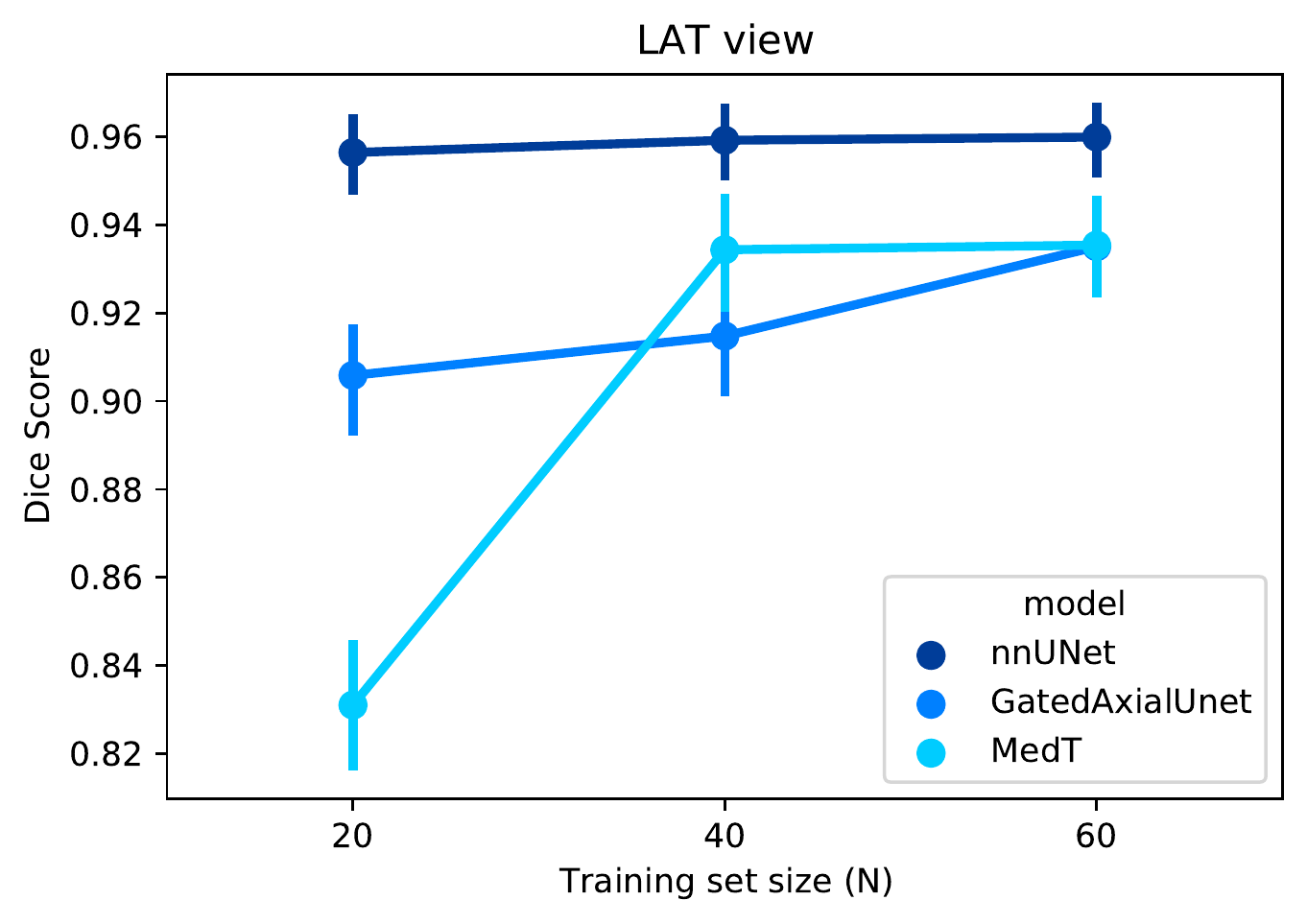}
        \caption{LAT Dice}
        \label{fig:LAT_dice}
    \end{subfigure}
    \vspace{0.2cm}
    \caption{Dice metric progression with incremental learning.}
    \label{fig:dice-jaccard-incremental}
\end{figure}

Figure \ref{fig:dice-jaccard-incremental} shows how model performance varied depending on the amount of data (20, 40 or 60 images) used to train the models (see Table \ref{tab:ap-lat-seg-metrics} for numerical results). nnU-Net provides greater stability than the other architectures, with good enough Dice (F1) metrics at low amounts of training data. Both MedT and GatedAxialUNet yielded expected results with incremental performance for both AP and LAT views. MedT required enough quantity of data to yield competitive results. In LAT images, this effect was perceived with greater emphasis.


Incremental learning showed, in general, significant improvement in performance for all architectures. Model performance increase was more noticeable in MedT and GatedAxialUNet. nnU-Net proved to have greater stability towards training data quantity variation, yielding promising results even with low training data availability. With only 20 images, nnU-Net performed similarly as with 60 images in both AP and LAT views.

\begin{table}[htb]
\centering
\resizebox{0.62\textwidth}{!}{%
\begin{tabular}{cccccc}
\hline
\multicolumn{6}{|c|}{\textbf{AP}} \\ \hline
\multicolumn{1}{|c|}{\textbf{N}} &
  \multicolumn{1}{c|}{\textbf{Metric}} &
  \textbf{DICE (\%)} &
  \textbf{PRC (\%)} &
  \textbf{RCL (\%)} &
  \multicolumn{1}{c|}{\textbf{ASD (px)}} \\ \hline
\multicolumn{1}{|c|}{\multirow{3}{*}{\textbf{20}}} &
  \multicolumn{1}{c|}{\textbf{nnUNet}} &
  96.47 ± 2.23 &
  97.4 ± 2.08 &
  95.67 ± 4.01 &
  \multicolumn{1}{c|}{2.15 ± 1.35} \\
\multicolumn{1}{|c|}{} &
  \multicolumn{1}{c|}{\textbf{GatedAxialUNet}} &
  91.83 ± 3.06 &
  97.33 ± 2.66 &
  87.04 ± 4.66 &
  \multicolumn{1}{c|}{4.75 ± 1.54} \\
\multicolumn{1}{|c|}{} &
  \multicolumn{1}{c|}{\textbf{MedT}} &
  86.78 ± 2.86 &
  87.86 ± 2.22 &
  85.84 ± 4.63 &
  \multicolumn{1}{c|}{15.23 ± 1.77} \\ \hline
\multicolumn{1}{|c|}{\multirow{3}{*}{\textbf{40}}} &
  \multicolumn{1}{c|}{\textbf{nnUNet}} &
  96.98 ± 1.96 &
  98.62 ± 1.17 &
  95.48 ± 3.79 &
  \multicolumn{1}{c|}{1.85 ± 1.27} \\
\multicolumn{1}{|c|}{} &
  \multicolumn{1}{c|}{\textbf{GatedAxialUNet}} &
  95.33 ± 3.01 &
  96.84 ± 2.51 &
  94.06 ± 5.34 &
  \multicolumn{1}{c|}{2.96 ± 1.65} \\
\multicolumn{1}{|c|}{} &
  \multicolumn{1}{c|}{\textbf{MedT}} &
  91.02 ± 2.3 &
  91.48 ± 1.63 &
  90.67 ± 4.22 &
  \multicolumn{1}{c|}{7.45 ± 2.29} \\ \hline
\multicolumn{1}{|c|}{\multirow{3}{*}{\textbf{60}}} &
  \multicolumn{1}{c|}{\textbf{nnUNet}} &
  97.09 ± 1.9 &
  99.39 ± 0.46 &
  94.96 ± 3.65 &
  \multicolumn{1}{c|}{1.56 ± 0.82} \\
\multicolumn{1}{|c|}{} &
  \multicolumn{1}{c|}{\textbf{GatedAxialUNet}} &
  94.53 ± 2.53 &
  97.95 ± 1.22 &
  91.47 ± 4.58 &
  \multicolumn{1}{c|}{3.28 ± 1.32} \\
\multicolumn{1}{|c|}{} &
  \multicolumn{1}{c|}{\textbf{MedT}} &
  93.54 ± 2.41 &
  99.33 ± 0.78 &
  88.47 ± 4.04 &
  \multicolumn{1}{c|}{4.17 ± 1.34} \\ \hline
\multicolumn{1}{l}{} &
  \multicolumn{1}{l}{} &
  \multicolumn{1}{l}{} &
  \multicolumn{1}{l}{} &
  \multicolumn{1}{l}{} &
  \multicolumn{1}{l}{} \\ \hline
\multicolumn{6}{|c|}{\textbf{LAT}} \\ \hline
\multicolumn{1}{|c|}{\textbf{N}} &
  \multicolumn{1}{c|}{\textbf{Metric}} &
  \textbf{DICE (\%)} &
  \textbf{PRC (\%)} &
  \textbf{RCL (\%)} &
  \multicolumn{1}{c|}{\textbf{ASD (px)}} \\ \hline
\multicolumn{1}{|c|}{\multirow{3}{*}{\textbf{20}}} &
  \multicolumn{1}{c|}{\textbf{nnUNet}} &
  95.18 ± 2.7 &
  97.21 ± 2.14 &
  93.32 ± 4.24 &
  \multicolumn{1}{c|}{4.28 ± 3.0} \\
\multicolumn{1}{|c|}{} &
  \multicolumn{1}{c|}{\textbf{GatedAxialUNet}} &
  90.58 ± 3.12 &
  95.43 ± 5.12 &
  86.54 ± 4.89 &
  \multicolumn{1}{c|}{7.55 ± 2.79} \\
\multicolumn{1}{|c|}{} &
  \multicolumn{1}{c|}{\textbf{MedT}} &
  83.1 ± 4.06 &
  83.84 ± 6.26 &
  82.67 ± 4.36 &
  \multicolumn{1}{c|}{25.11 ± 1.46} \\ \hline
\multicolumn{1}{|c|}{\multirow{3}{*}{\textbf{40}}} &
  \multicolumn{1}{c|}{\textbf{nnUNet}} &
  95.26 ± 2.66 &
  97.98 ± 1.79 &
  92.78 ± 4.41 &
  \multicolumn{1}{c|}{3.36 ± 1.35} \\
\multicolumn{1}{|c|}{} &
  \multicolumn{1}{c|}{\textbf{GatedAxialUNet}} &
  91.48 ± 3.42 &
  95.49 ± 4.04 &
  88.13 ± 5.93 &
  \multicolumn{1}{c|}{6.91 ± 2.87} \\
\multicolumn{1}{|c|}{} &
  \multicolumn{1}{c|}{\textbf{MedT}} &
  93.43 ± 3.5 &
  94.14 ± 4.83 &
  92.95 ± 4.31 &
  \multicolumn{1}{c|}{6.38 ± 3.33} \\ \hline
\multicolumn{1}{|c|}{\multirow{3}{*}{\textbf{60}}} &
  \multicolumn{1}{c|}{\textbf{nnUNet}} &
  95.38 ± 2.53 &
  98.17 ± 2.0 &
  92.84 ± 4.2 &
  \multicolumn{1}{c|}{3.26 ± 1.4} \\
\multicolumn{1}{|c|}{} &
  \multicolumn{1}{c|}{\textbf{GatedAxialUNet}} &
  93.5 ± 1.97 &
  95.94 ± 3.38 &
  91.32 ± 3.02 &
  \multicolumn{1}{c|}{5.29 ± 1.82} \\
\multicolumn{1}{|c|}{} &
  \multicolumn{1}{c|}{\textbf{MedT}} &
  93.54 ± 2.93 &
  96.7 ± 2.97 &
  90.74 ± 4.7 &
  \multicolumn{1}{c|}{6.15 ± 2.76} \\ \hline
\end{tabular}%
}
\vspace{0.32cm}
\caption{Metrics for each of the configurations trained on both AP (top) and LAT (bottom) images. Results are obtained from testing each model on the independent CISM test set (AP and LAT) following an incremental learning approach. N refers to the number of images with which each model was trained. DICE: Dice, PRC: Precision, RCL: Recall, ASD: Average Surface Distance.}
\label{tab:ap-lat-seg-metrics}
\end{table}

\subsubsection{Results comparison}

The most stable and best performing architecture was nnU-Net. Nonetheless, GatedAxialUNet and MedT also yielded good performance results, even carrying out a more efficient training process than nnU-Net (time delayed during the training process was drastically reduced with MedT and GatedAxialUNet). However, performance metrics provided by these last two models did not reach as high values as nnU-Net did.

\begin{figure}[htb]
    \centering
    \includegraphics[width=0.6\textwidth]{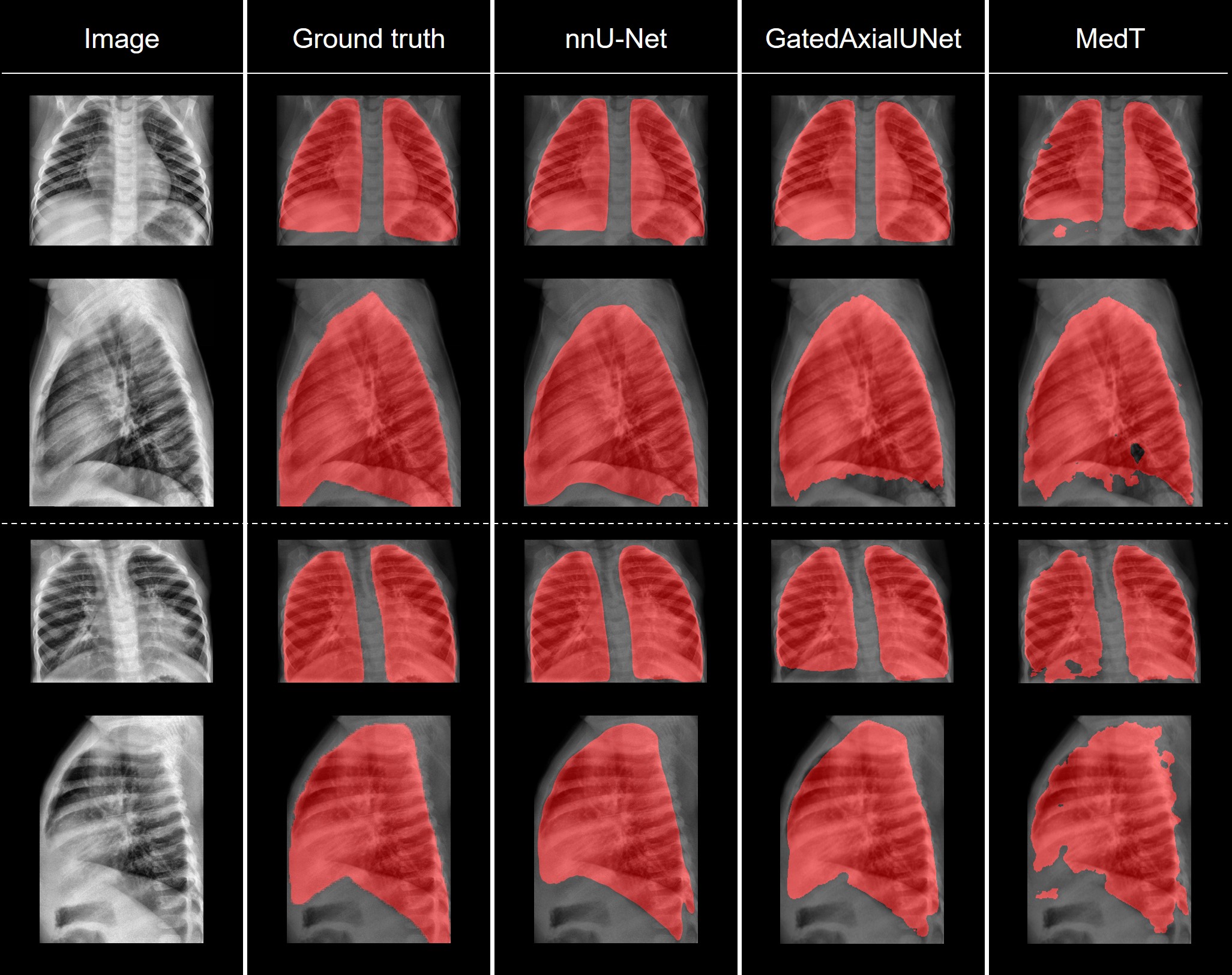}
    \vspace{0.3cm}
    \caption{Visual comparison of the predictions obtained from each of the models in a 29-month-old non-TB case (up) and an 8-month-old TB-compatible case (down). These cases belong to the independent CISM test set.}
    \label{fig:seg_comparison}
\end{figure}

Figure \ref{fig:seg_comparison} shows a visual comparison of the predictions obtained from each of the models in a non-TB case and a TB-compatible case from the independent test set.

Thus, nnU-Net demonstrated greater capacity in segmenting lungs in both AP and LAT views, even when fewer images were used for training. Nonetheless, training and inference times were much shorter in GatedAxialUNet and MedT.

\subsection{Automatic LAT Orientation Correction}

The custom ResNet model implemented for this step provided an accuracy of 1.00 in the test set, correctly detecting if the vertebral column was located at the right or the left in the LAT-view image. As expected, neither false positives nor false negatives were detected among the test set predictions, as the problem was relatively simple for the network, although necessary to provide the system with greater robustness.

\subsection{Standardized Template and Automatic Region Extraction}

\begin{figure}[htb]
    \centering
    \includegraphics[width=0.75\textwidth]{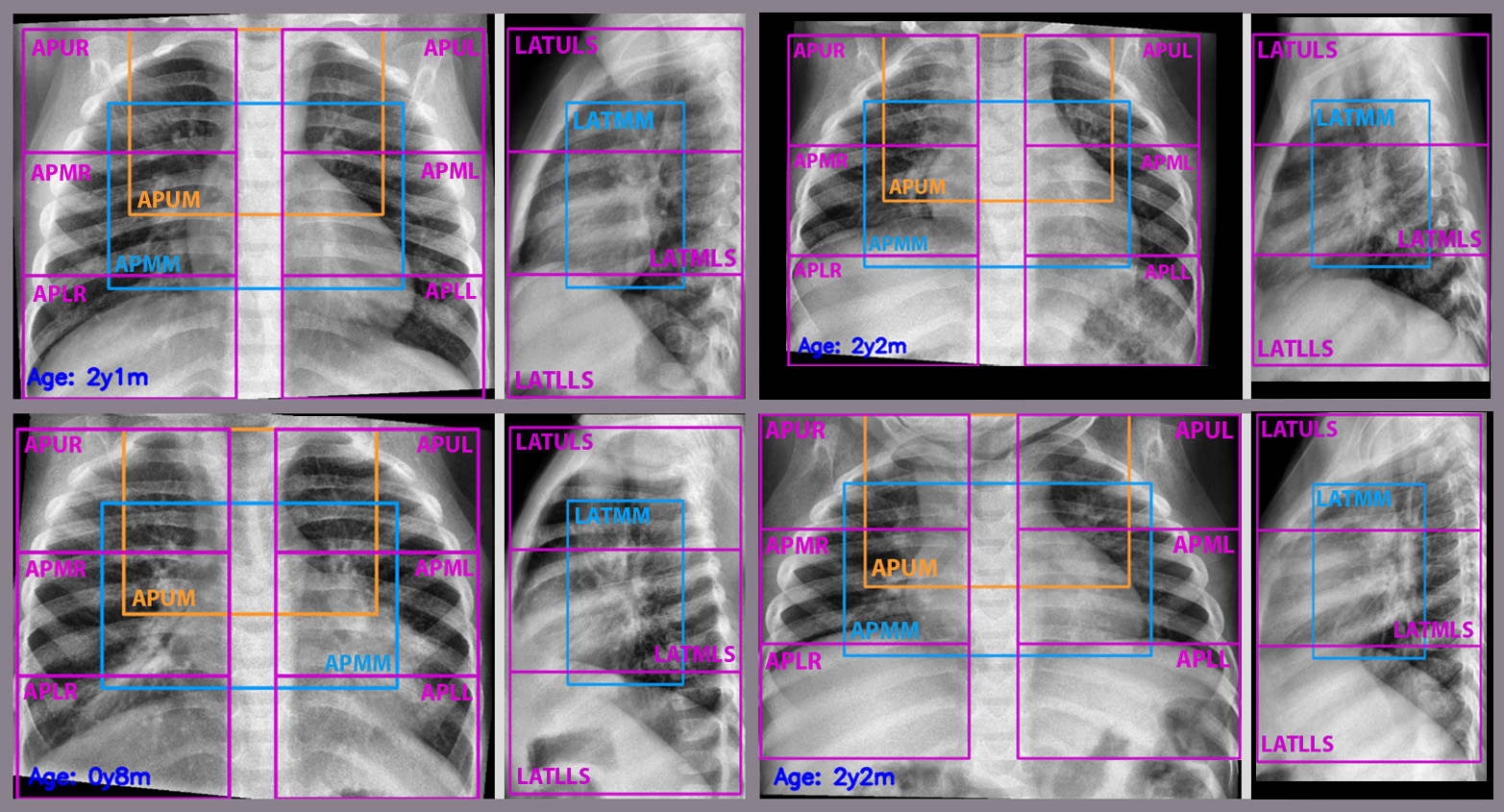}
    \vspace{0.3cm}
    \caption{Automatic region extraction in two healthy (up) and two TB-compatible cases (bottom). Regions extracted are labelled according to those defined in section \textit{\ref{ssec:auto_region_extraction_methods}}.}
    \label{fig:montage-region-extraction}
\end{figure}

Finally, the template construction and regional partition was then tested on the independent CISM test set. As input, the predictions used for this final step corresponded to the output of the nnU-Net model trained with 60 images, which demonstrated to have the best performance on the lung segmentation task. An expert radiologist (RSJ, from author list) performed a visual validation of the results. From the 60 CISM AP and LAT test images corresponding to the 30 CISM test cases, 54 were marked as correct, in 5 images minimal corrections (no substantial difference would be perceived in further region-linked TB finding assessment) were suggested, and only in 1 image severe corrections (substantial difference would be perceived in further assessment) were reported. Figure \ref{fig:montage-region-extraction} presents four randomly selected cases of the test set, showing how these regions are extracted in different scenarios, in cases of different age.

\section{Conclusions}
\label{sec:conclusions}

In this paper, we have proposed a multi-view deep learning-based pipeline which automatically extracts lung and mediastinal regions of interest from pediatric CXR images, based on a previously proposed standard template. This standard template and its partitions can be used for further analysis to confirm TB findings presence and severity assessment given a pediatric CXR, where TB assessment entails a challenging task. The proposed system lays the groundwork for automatic approaches that may reduce the high clinical burden when assessing pediatric TB, especially in countries with low resources and high prevalence of TB.

\acknowledgments 

This work was supported by H2020-MSCA-RISE-2018 INNOVA4TB (EU) project (ID 823854) and ADVANCE-TB Cost Action (EU) project (ID CA21164). DCM's PhD fellowship was supported by Universidad Politécnica de Madrid.

\bibliography{references} 
\bibliographystyle{spiebib_3etal} 

\end{document}